\documentclass{ecai}
\usepackage{graphicx}
\usepackage{times}
\usepackage{latexsym}

\usepackage{microtype}
\usepackage{booktabs} 
\usepackage{amsmath,amssymb}
\usepackage{enumerate}
\usepackage{enumitem}
\usepackage{multirow}
\usepackage{url}

\newcommand{\remove}[1]{}

\usepackage{glossaries}

\newacronym{BOW}{BOW}{Bag-of-Words}
\newacronym{TFIDF}{TFIDF}{Term Frequency - Inverse Document Frequency}
\newacronym{LSI}{LSI}{Latent Semantic Indexing}
\newacronym{LDA}{LDA}{Latent Dirichlet Allocation}
\newacronym{WMD}{WMD}{Word Mover\rq s Distance}
\newacronym{RWMD}{RWMD}{Relaxed Word Mover\rq s Distance}
\newacronym{WCD}{WCD}{Word Centroid Distance}
\newacronym{S-WMD}{S-WMD}{Supervised Word Mover\rq s Distance}
\newacronym{EMD}{EMD}{Earth Mover\rq s Distance}
\newacronym{SVD}{SVD}{Singular Value Decomposition}

\begin{document}

\title{Speeding up Word Mover's Distance and its variants via properties
of distances between embeddings}

\author{
Matheus Werner \and Eduardo Laber
\institute{Departamento de Inform{\'a}tica, PUC-Rio, Rio de Janeiro, Brazil, email: \{mwerner, laber\}@inf.puc-rio.br} }

\maketitle
\bibliographystyle{ecai}

\begin{abstract}
The Word Mover's Distance (WMD) proposed by Kusner et al. is a distance between documents that takes advantage of semantic relations among words that are captured by their embeddings. This distance proved to be quite effective, obtaining state-of-art error rates for classification tasks, but is also impracticable for large collections/documents due to its computational complexity. For circumventing this problem, variants of WMD have been proposed. Among them, Relaxed Word Mover's Distance (RWMD) is one of the most successful due to its simplicity, effectiveness, and also because of its fast implementations.

Relying on assumptions that are supported by empirical properties of the distances between embeddings, we propose an approach to speed up both WMD and RWMD. Experiments over 10 datasets suggest that our approach leads to a significant speed-up in document classification tasks while maintaining the same error rates.
\end{abstract}

\section{INTRODUCTION}

Document comparison is a fundamental step in several applications such as recommendation, clustering, search, and categorization. In its simplest version, this task consists of computing the distance between a single pair of documents.

The document representation is an essential factor in the definition of a distance. Arguably, the most employed document representations due to its simplicity and good results are the  Bag-of-Words (BOW) and the Term Frequency - Inverse Document Frequency (TF-IDF). These representations are based on word counts, and so they may lose information that is relevant for some applications, such as the ordering among words in a document, co-occurrence, and semantic relations between different words.  Thus, richer representations that take into account some of this information have been proposed \cite{shannon19481948,dumais1988using, blei2003latent}.

Up to a few years ago, semantic relations were barely used because there was no adequate methodology of how to obtain them. Consequently, researchers eventually decided to use ontologies as a way to mitigate this issue \cite{hotho2003ontologies}, although this makes applications dependent on an external knowledge base. 
This scenario changed with the emergence of Word2Vec \cite{mikolov2013efficient, mikolov2013distributed} and its variants \cite{pennington2014glove}, a class of methods that allow us to efficiently identify the relationship between words and embed them into vectors, called word embeddings. As a result, researchers have been searching for ways to use these embeddings to refine existing models in the literature. The results of these efforts can already be seen in works such as \cite{le2014distributed,kusner2015word,das2015gaussian,li2016topic}, and indeed, improvements are obtained.

In particular, Kusner et al. \cite{kusner2015word} propose the Word Mover\rq s Distance (WMD), an application of the classic Earth Mover\rq s Distance (EMD) \cite{rubner1998metric} for the domain of documents that takes advantage of the semantic relations captured by the embeddings associated with their words. The idea is to compute the minimum cost required to transform one document representation into another by using the distance between embeddings as the cost of transforming words. In fact,  the distance is given by the cost of an optimal solution of a transportation problem defined on a complete bipartite graph where the nodes correspond to the distinct words of the documents, and the edge costs are the distance between embeddings. In the same paper, they show that this approach obtained outstanding results on document classification tasks, outperforming many competitors. The major drawback of WMD, however, is its high computational cost since solving the transportation problem in a complete bipartite graph is costly, requiring super cubic time.

Since the proposal of WMD, there has been a considerable amount of research focusing on  improving its performance while
keeping its effectiveness \cite{kusner2015word, atasu2017linear, wu2018word, atasu2019linear}.
The Relaxed Word Mover's Distance (RWMD) \cite{kusner2015word,atasu2017linear,atasu2019linear}, due to its  simplicity and speed, is arguably
one of the most successful outcomes of this research effort.
In fact, experiments reported in the literature show that it achieves quality (test error) competitive with those obtained by  WMD with the advantage of being much faster. However, despite its good performance, further improvements are relevant because there are applications in which this kind of distance needs to be calculated very quickly.

Motivated by this scenario, we focus on developing an approach to derive distances that are as effective as WMD and its variants with the advantage of allowing a  faster computation.

\subsection{Our Contributions}

To achieve this goal, in contrast to other approaches available,  we explore the properties of the application domain, more specifically the distribution of distances among word embeddings. Our  key observation is that one can assume, without incurring a significant loss, that the set of distances between word embeddings is split into two sets: the set of distances between related words and the set of distances between non-related words, with the distances in the latter  having the same value. We show that   this assumption, which is supported by empirical data, can be used  to: (i) obtain a more compact formulation for the transportation problem that is used to calculate WMD and its variants and (ii) dramatically reduce the memory required to cache the distances between embeddings, which  is essential for the fast computation of RWMD for large vocabularies since the evaluation of the distance between a pair of words requires hundreds of operations for typical sizes of embeddings.

By relying on the previous observation, we propose a simple approach  for speeding up WMD and distances with a similar flavour.
More concretely, we show how to derive new distances between documents by applying our approach to both WMD and RWMD. The time and space complexities required to compute these distances depend on a parameter $r$ that has to do with the number of related words. 
This parameter can be set to a small value which leads to  complexity improvements over WMD and RWMD.
In addition,  experiments  executed over 10 datasets, for two distinct  tasks, suggest that these distances yield to test errors as good as those obtained by WMD/RWMD, with a significant gain in terms of execution time.
Indeed, with regards to  efficient implementations of RWMD, we obtained an average speed-up of almost 5 times for one task and 15 times for the other.

\subsection{Related Work}
Our work is closely related to some approaches that have been proposed to circumvent the high computational cost of WMD \cite{kusner2015word, atasu2017linear, wu2018word, atasu2019linear}.

Kusner et al. \cite{kusner2015word} propose the Relaxed Word Mover\rq s Distance (RWMD), a distance that is defined over a relaxation of the transportation problem in which some constraints are dropped. Given the  distance matrix between the words embeddings of documents  $D$ and $D'$, the RWMD can be calculated  in  $O (|D|\cdot |D'|) $ time, where $|D|$ and $|D'|$ are  the number of distinct words of $D$ and $D'$, respectively. Thus, the bottleneck of RWMD is to compute the distance matrix  which costs $ O (|D|\cdot |D'| \cdot d)$ time,  where $d$ is the dimension of the word embeddings space. Such cost can be prevented by caching the $ O(n^2)$ distances between all the $n$ words of the vocabulary, an approach that could be prohibitive for large $n$. Experiments from \cite{kusner2015word} shows that RWMD achieves test error competitive with  WMD for document classification tasks while incurring a  lower computational cost, even without using cache.

Atasu et al. \cite{atasu2017linear} show how to compute RWMD for any two documents $D$ and $D'$ from a collection ${\cal C}$ in $ O(|D|+|D'|)$ time. To achieve this running time, they need to  pre-compute and store the distance of word $w$ to the nearest word in document $D$, for each  $w$  in the vocabulary and each  $D$ in the collection. Thus, it  consumes $ O(n |{\cal C}|)$ memory, where $|{\cal C}|$ is the number of documents in  ${\cal C}$, which may be infeasible for large collections. Furthermore, this linear time complexity does not hold for dynamic collections since the method  requires $O(|D^{new}| \cdot n \cdot d)$ preprocessing time before calculating the RWMD from a new document $D^{new}$ to some document $D$.
Further work from Atasu et al. \cite{atasu2019linear} 
discusses a limitation of RWMD for documents that share many words 
and then proposes a family of 
variants of RWMD that better
address this scenario.

In a broader scope, our work is also related to some proposals to speed up EMD \cite{pele2009fast} and approximate solutions for transportation problems in general \cite{cuturi2013sinkhorn, genevay2016stochastic, peyre2019computational}.

Pele et al. \cite{pele2009fast} present an optimized solution of the EMD for instances in which the costs of the edges satisfies  certain properties that are motivated by the way human perceive distances. The optimization introduced by this approach consists of reducing the number of edges in the transportation network and, as a consequence, the running time. This work resembles ours in the sense that both optimize the time complexity to solve the transportation problem by taking into account how the costs behave in the domains under consideration.

Cuturi et al. \cite{cuturi2013sinkhorn} use an entropic regularization term to smooth out the transportation problem so that it can be solved much faster via  Sinkhorn\rq s matrix scaling algorithm. This algorithm has $O(|D| \cdot |D'|)$  empirical time according to \cite{cuturi2013sinkhorn} and it was used  in a supervised version of WMD \cite{huang2016supervised}. As RWMD, this method needs an $O(n^2)$ space cache in order to prevent the $O(|D| \cdot |D'| \cdot d)$ time  required to compute the distances between the words in $D$ and $D'$.

\subsection{Paper Organization}
The paper is organized as follows. In Section \ref{sec:background}, we introduce our notation and discuss some background that is important to the understanding of our work. In the next section, we develop our approach. In Section  \ref{sec:results}, we present  our experimental study comparing the new distance with WMD and RWMD both in terms of test error and computational performance. Finally, in Section \ref{sec:conclusions}, we present our conclusion.

\section{BACKGROUND}
\label{sec:background}
In this section, we introduce some notation and  explain some
 concepts
that are important to understand our work.

\subsection{Notation}
We assume that we have a vocabulary of $n$ words $\{1,\ldots,n\}$ and a collection of documents. 
Throughout the text, we need to refer to arbitrary documents $D$ and $D'$
to explain existing distances and the new ones that we propose.
Hence, unless otherwise stated,  we assume that  the set of distinct words of $D$ and $D'$ are, respectively, $\{w_1,\ldots,w_{|D|}\}$ and $\{w'_1,\ldots,w'_{|D'|}\}$.
Note that  $|D|$ (resp. $|D'|$) is the number of distinct words of document $D$ (resp. $D'$).  Moreover, we use $D_i$  to denote the normalized frequency of
$w_i$, that is, the number of occurrences of
$w_i$ in $D$ over the total
number of words in $D$. We use $D'_j$ to refer 
to the normalized frequency of $w'_j$ analogously.
Note that $\sum_i D_i = \sum_j D'_j=1$.

We use ${\bf x}(i)$
to denote the embedding of a word $i$  in a vector space of dimension $d$ and we use  $c(i,j)$ to denote the Euclidean distance between the embeddings of words $i$ and  $j$,
that is,  $c(i,j)=\|{\bf x}(i)-{\bf x}(j)\|_2$.

To make sure that our notation is clearly understood, we
present a simple  example involving the following
documents:

\begin{equation*}
\begin{array}{ll@{}ll}
\textbf{$D$: } & \text{John likes algorithms. Mary likes algorithms too.} \\
\textbf{$D'$: } & \text{John also likes data structures.}
\end{array}
\end{equation*}

Ignoring the stopwords, and assuming
that $D$ and $D'$  are the only doc's in our 
collection, we have the following vocabulary
\begin{equation*}
\begin{array}{lll@{}lll}
\{ 1:\text{``John''}, & 2:\text{``likes''}, & 3:\text{``algorithms''}, & 4:\text{``Mary''}, \\ 
\ \ 5:\text{``too''}, & 6:\text{``also''}, & 7:\text{``data''}, & 8:\text{``structures''} \} \\
\end{array}
\end{equation*}
Hence,  the set of distinct words of $D$ and $D'$ are, respectively, 
$ \{w_1=1, w_2=2, w_3=3, w_4=4, w_5=5\}$ and
 $\{w'_1=1, w'_2=6, w'_3=2, w'_4=7, w'_5=8\}$. 
 Finally, the normalized \glsentrylong{BOW} representations of $D$ and $D'$ are, respectively,
\begin{equation*}
\begin{array}{ll@{}ll}
\textbf{$D$: }  & \{D_1, D_2, D_3, D_4, D_5\} = \{1/7, 2/7, 2/7, 1/7,1/7\} \\
\textbf{$D'$: } & \{D'_1, D'_2, D'_3, D'_4, D'_5\} = \{1/5, 1/5, 1/5, 1/5, 1/5\}. \\
\end{array}
\end{equation*}

\subsection{\glsentrylong{WMD}}
\label{ssec:WMD}

By exploring the behaviour of the Word Embeddings, Kusner et al. \cite{kusner2015word} define the distance between two documents as the minimum cost of converting the words of one document into the words of the other, where the cost $c(w_i,w'_j)$ of transforming the word $w_i$ into word $w'_j$ is given by the distance between the word embeddings of $w_i$ and $w'_j$.

\vskip 0.5in
Formally, the WMD between documents $D$ and $D'$ is defined as the value of the optimal solution of the following transportation problem:

\begin{align}
\text{min}  & \displaystyle\sum\limits_{i=1}^{|D|}
\displaystyle\sum\limits_{j=1}^{|D'|}  c(w_i,w'_j) T_{ij} & \label{eq:wmd-formulation-0} \\
\text{s.t.:}& \displaystyle\sum\limits_{j=1}^{|D'|}   T_{ij} = D_i   & \quad \quad \forall i \in \{1 ,\dots, |D| \} \label{eq:wmd-(A)} \\
 & \displaystyle\sum\limits_{i=1}^{|D|}   T_{ij} = D'_j  & \quad \quad \forall j \in \{1 ,\dots, |D'| \} \label{eq:wmd-(B)}   \\
& T_{ij} \ge 0 & \quad \text{for all } \quad i,j
\label{eq:wmd-formulation}         
\end{align}

In the above formulation, $T$ is the flow matrix. The variable  $T_{ij}$ gives the  amount of word $w_i$ that is transformed into word $w'_j$. 
The equation (\ref{eq:wmd-(A)}), for a fixed $i$,  assures that each unit of word $w_i$ is transformed into a unit of a word in $D'$ while  Equation (\ref{eq:wmd-(B)}), for each $j$, assures that the total units of words in  $D$ transformed into $w'_j$ is $D'_j$.

The WMD, although well founded, suffers from efficiency problems since solving  the transportation problem on a complete bipartite graph is costly, requiring super cubic time using the best known minimum cost flow algorithms \cite{pele2009fast}.

\subsection{Relaxed Word Mover\rq s Distance}
\label{ssec:rwmd}

To overcome the high computational cost of
solving the transportation problem,
 Kusner et al. \cite{kusner2015word} propose
 the RWMD, a variation of WMD whose computation relies on optimally solving  
 relaxations of the transportation problem.
These  relaxations are obtained by either ignoring  the set of constraints 
 (\ref{eq:wmd-(A)}) or the set
 (\ref{eq:wmd-(B)}). 
 In fact, the RMWD between $D$ and $D'$ 
can be calculated by evaluating the  expression
\begin{equation}
\max \left \{ \sum_{i=1}^{|D|} D_i  \min_{ 1 \le j \le |D'|} c(w_i,w'_j) ,  
\sum_{j=1}^{|D'|} D'_j  \min_{1 \le i \le |D|} c(w_i,w'_j) \right \}
\label{eq:rwmd}
\end{equation}
where the left and the right terms in the maximum  are the optimum values of the relaxations that ignore constraints (\ref{eq:wmd-(B)}) and (\ref{eq:wmd-(A)}), respectively.

By examining the above equation we conclude that, given the costs $c(w_i,w'_j)$'s, RWMD can be calculated for a pair of documents $D$ and $D'$ in $O(|D| \times |D'|)$ time, which is a significant improvement over WMD. Therefore,  RWMD's bottleneck is the computation of the costs $c(w_i,w'_j)$'s since it requires $O(|D| \cdot |D'| \cdot d)$ time.

We also note that by performing a simple preprocessing step before applying equation (\ref{eq:rwmd}) we can obtain a tighter relaxation of WMD that corresponds to the OMR distance proposed in \cite{atasu2019linear}. The motivation is better handling cases in which $D$ and $D'$ share many words. 

The preprocessing
consists of first identifying pairs of words $(w_i,w'_j)$, with $w_i=w'_j$. Then, for each of these pairs
we do the following:
(i) we replace $D_i$ with its excess $\max \{D_i-D'_j,0\}$
and, if $D_i \le D'_j$, we remove index $i$ from the range where the
minimum iterates in the right term of the max; (ii) similarly,
we replace $D'_j$ with its excess $\max \{D'_j-D_i,0\}$
and, if $D'_j \le D_i$, we remove index $j$ from the range where the
minimum iterates in the left term of the max. 
The impact of this preprocessing is 
associating in equation (\ref{eq:rwmd}) 
the excesses of $D_i$ and $D'_j$
with the second closest word to $w_i$ and $w'_j$, respectively.

\subsubsection{RWMD in linear time}
\label{ssec:LC-RWMD}
In \cite{atasu2017linear}, it is proposed 
 an implementation that computes the RWMD  between two documents $D$ and $D'$ in 
$O(|D|+|D'|)$ time, improving upon
the $O(|D| \cdot |D| \cdot d )$ time required by the original 
proposal. 
This improvement, however, comes at the expense of some potentially costly  preprocessing.

To explain the implementation, denoted here by RWMD(L),
 let ${\cal C}$ be a collection of documents and
  let ${\cal C}_i(j)$ be the
word, among those  in the $i$-th document of ${\cal C}$, that is closest to some given word $j$ in the vocabulary.  RWMD(L)  builds,
at the preprocessing phase,
a matrix   $M$ with $|{\cal C} |$ rows and $n$ columns, where the
 entry $M_{ij}$ stores the distance between ${\cal C}_i(j)$
and word $j$.  To  fill the row of $M$ 
associated with document $D$ we have to pay $O(n \cdot |D| \cdot d )$ time.

Having the matrix $M$ available, it
is possible to compute the RWMD between documents $D$ and $D'$ 
in $O(|D|+|D'|)$ time. The reason is
that the terms
$min_{j} c(w_i,w'_j)$
and $min_{i} c(w_i,w'_j)$ of (\ref{eq:rwmd})
can be computed in $O(1)$ time.
The former is obtained by accessing the entry $M_{r' ,i}$, where $r'$ is the row corresponding to document $D'$,
while for the latter we need to access
the entry $M_{r,j}$, where $r$ is the row corresponding to document $D$.

In addition to the time required to build
matrix $M$, another potential problem of  RWMD(L)  is its space requirement since the matrix
$M$ can be prohibitively large  when either the vocabulary or the 
collection of documents is huge.  	

We note that in order to properly use  the preprocesssing
for equation (\ref{eq:rwmd}) described
in the previous section, 
we should store in matrix $M$ the distance of $w$ to its 
second closest word among those that belong $D$, when $w \in D$.

\section{AN APPROACH FOR SPEEDING UP WMD AND ITS VARIANTS}
\label{sec:efficient-methods}

For large vocabularies the methods discussed
in the previous section may have to cope with an enormous amount 
of distances between embeddings, which
may  be a problem either in terms
of memory requirements or in terms of running time.
In fact, for applications (e.g.  document classification via $k$-NN) that  use
the distance between the same  pair of embeddings several
times  caching turns out to be crucial for achieving
computational efficiency. However, when the vocabulary is
large, caching becomes prohibitive.

In this section, we show that we can significantly attenuate this problem by taking into account the distribution of distances among word embeddings.

\subsection{On the distances between word embeddings}
\label{sec:distance-assumptions}

Here we discuss the assumptions in which our approach
and the new distances, derived from it, rely on. For that, we present examples of distances between word embeddings. 
These embeddings, used here and in the next sections, were made available by Google \footnote{https://code.google.com/archive/p/word2vec/}. To obtain them, they trained the vectors with $d=300$ using the Word2Vec template of Word Embeddings \cite {mikolov2013distributed} on top of a Google News document base containing altogether about 100 billion words and 3 million tokens. We also refer to some datasets that will be detailed in our experimental section.

From a semantic perspective, it is reasonable to consider that words are closely related to only a few other words in general. As the word embeddings were designed to simulate semantic relations, it is expected that they present a similar behaviour; that is, each vector should be close to a few other vectors and far away from the remaining ones. 

As an example, if the words are ranked according to their distances to the embedding corresponding to ``cat'' one should expect ``dog'' and ``rabbit'' preceding both ``moon'' and ``guitar''. However, it is not clear whether ``moon'' or ``guitar'' comes first in the ranking since neither of them has an obvious relation with ``cat''.  

\begin{figure}[ht]
\centerline{\includegraphics[width=0.8\columnwidth]{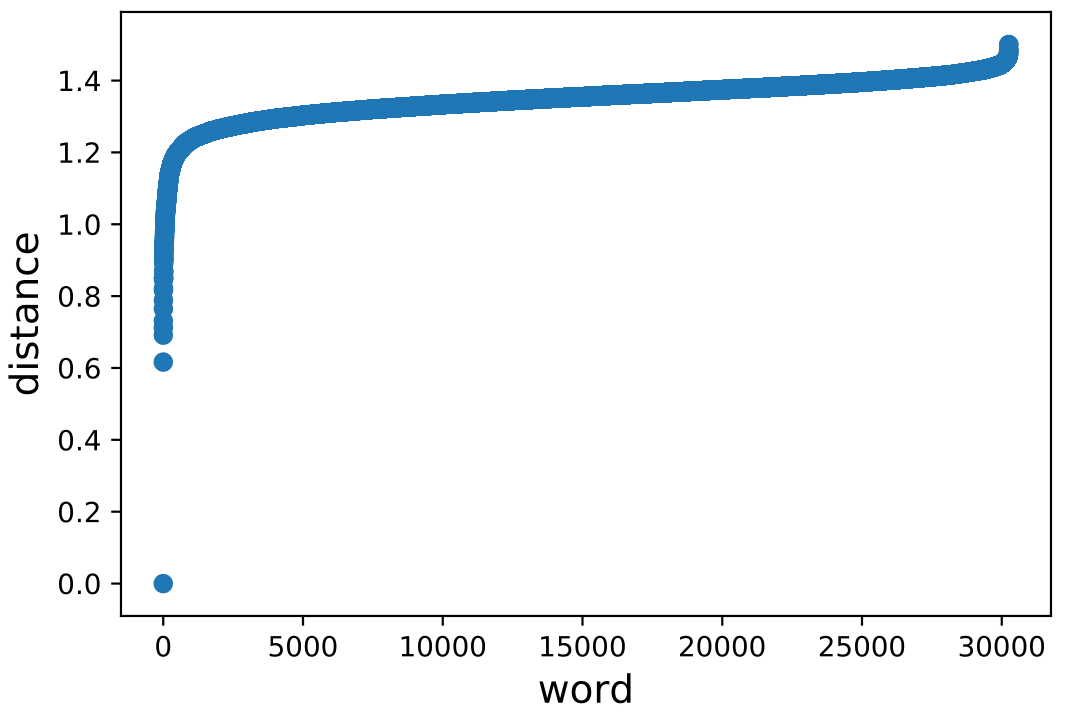}}
\caption{Distances from embeddings of all words in the vocabulary of Amazon dataset to the word ``cat''.}
\label{fig:word-embeddings}
\end{figure}

Figure \ref{fig:word-embeddings} illustrates this behaviour by displaying the distances between the embedding for ``cat'' and the embeddings from the words of the Amazon dataset sorted by increasing order of distance. We note that there are few words with small distances while the vast majority has distance concentrated in the range $[1.2,1.4]$.

\begin{figure}[ht]
\centerline{\includegraphics[width=0.8\columnwidth]{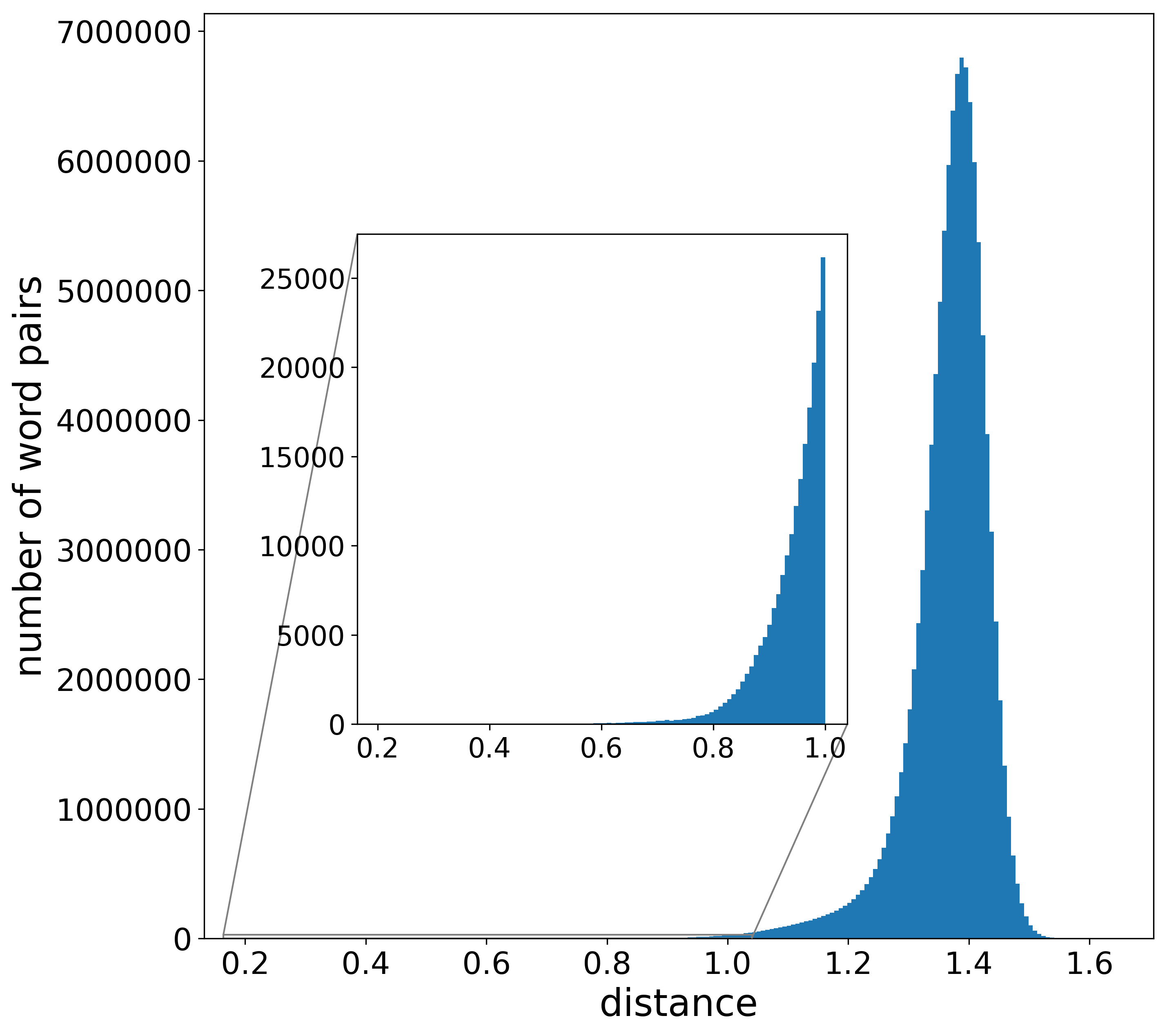}}
\caption{Distribution of the distances between all words in the vocabulary of Reuters dataset.}
\label{fig:distance-distribution}
\end{figure}

For checking whether this behaviour persists for other words, we computed all the distances between embeddings from the words of the Reuters dataset. 
Figure \ref{fig:distance-distribution} shows the distribution of these distances clustered in bins for better visualization. Once again, we observe a high concentration of the distances around the interval $[1.2, 1.4]$, behaving similarly to a Normal distribution.

Based on this discussion, we make the following assumptions:

\begin{enumerate}[label=(\roman*)] 
	\item Given a word $w$, the remaining words can be split into two groups: {\tt RELATED}($w$) and {\tt UNRELATED}($w$), with the former (latter) containing the words related (unrelated) with $w$;
	\item The distances from every word in {\tt UNRELATED}($w$) to $w$, for every $w$, is the same ``large'' value  $c_{max}$.
\end{enumerate}

Although the number of related words may  vary
according to the word of reference, in order to make our approach simpler
and thus, more practical, we assume that all words have the same number of related words and we use $r$ to denote this number. The value of $r$ can either  be set 
manually  or automatically  estimated using a training set as we discuss in the experimental section.

\subsection{Algorithms exploiting distance assumptions}

Algorithms to compute distances  with the flavour of WMD can benefit from our assumptions because, by using them, they just need to handle a much smaller set of distances between embeddings, that is, the set of distances between related words. 
As a result, caching distances  becomes feasible even for large vocabularies, which prevent these methods of calculating the distance between the same pair of words more than once.
In addition, the transportation problem in which WMD and related distances as RWMD rely on can be solved in a sparse bipartite graph rather than on a complete bipartite graph.

In the next subsections we discuss how WMD and RWMD can be adapted to make use of our assumptions. These adaptations lead to  new distances between documents, namely Rel-WMD and Rel-RWMD. We start with the  explanation of a preprocessing phase that is required to calculate these new distances.

\subsubsection{Preprocessing Phase}
\label{sec:preprocessing}
In this phase, we build a structure (cache) $C$ that stores for each word $w$, from a vocabulary of $n$ words, the $r$ closest words to $w$ as well as its distances to $w$.

Choose a word $i$ in the vocabulary. The procedure computes  its Euclidean distance $\|{\bf x}(i)-{\bf x}(j)\|_2$ to every other word $j$ and add these distances, as well as the corresponding words, to  a list  $L_i$. 
Next, it selects the $r$ words that are closest to $i$ in $L_i$ and adds them, with their distances, to cache $C$. This selection can be performed in expected linear time  using {\tt QuickSelect} \cite{hoare1961algorithm}. 
The distances that were not included in $C$ are then  added to a global accumulator $A$ with the goal of calculating $c_{max}$. This procedure is repeated for every word $i$ in the vocabulary and the value $c_{max}$ is given by the average of all values added to $A$. 

The cache $C$ requires $O(n \cdot r)$ space and its construction requires $O(n^2 \cdot d )$ time, where the term $d$ is due
to the time required to compute the distance
between a pair of embeddings.

For large vocabularies the construction of the cache $C$,
as  above described, may be costly due to the  $O(n^2 \cdot d )$ time complexity.
This construction, however,  can be optimized by clustering the embeddings 
and then considering only words in the same cluster to
find the related words.  

We discuss
this approach using the traditional 
$k$-means clustering algorithm \cite{arthur2007k}.
On the one hand, this algorithm  allows the
user to define the number of clusters $k$ and it  performs $O(n  \cdot k \cdot d \cdot I)$ operations  to cluster $n$ points in $R^d$ into $k$ clusters, where $I$
is the maximum  number of iterations allowed. On the other hand, if the $n$ embeddings are uniformly distributed among
$k$ clusters then the construction of cache $C$ requires  $O((n/k)^2 \cdot d)$ time per cluster which implies on  $O(n^2 \cdot d/k)$ overall time.
Hence, let $f(k)= n \cdot d \cdot  k \cdot  I + (n^2 \cdot d/k)$ be an estimation of the running time  required to execute  $k$-means and then the
construction of cache $C$. By minimizing $f(k)$ we get $k= \sqrt{n/I}$. Thus, if $n$ is large, 
 in order to speed up the preprocessing phase,  we  run $k$-means algorithm with $k = \sqrt{n/I}$, 
before building the cache $C$.

\subsubsection{Related Word Mover\rq s Distance}
\label{ssec:rel-wmd}

The Related Word Mover\rq s Distance (Rel-WMD) between $D$ and $D'$ is defined as the optimum value of the transportation problem given by equations (\ref{eq:wmd-formulation-0})-(\ref{eq:wmd-formulation}), where the costs of the edges are as follows:

\begin{equation} 
 c(w_i,w'_j)= \begin{cases} 
	0, & \mbox{ if } w_i=w'_j \\
	\|{\bf x}(w_i)-{\bf x}(w'_j)\|_2, & \mbox{ if } (w_i,w'_j) \in C \\
	c_{max}, & \mbox{otherwise}
\end{cases} 
\label{eq:costs-wmd}
\end{equation}

For small values of parameter $r$ many costs are equal to $c_{max}$.
In this case,  it is possible to replace the formulation given by  (\ref{eq:wmd-formulation-0})-(\ref{eq:wmd-formulation}) with an equivalent and more compact one.
This new formulation is  given by (\ref{eq:rel-wmd-1})-(\ref{eq:rel-wmd-4}) and its key idea is using variable $T_{i,t}$ to represent the number of units of word $w_i$ that is transformed into words that are at a distance $c_{max}$ from $w_i$. Thus, the single variable $T_{i,t}$ replaces all variable $T_{i,j}$ in the original formulation for which $(w_i,w'_j)$ does not belong to cache $C$. 
Similarly $T_{t,j}$ represents the number of units transformed into $w'_j$  from words that are at a distance $c_{max}$ of word $w'_j$.
The underlying graph of this new formulation is much sparser (for small values of $r$) so that the transportation problem can be solved significantly faster.

\begin{align}
\label{eq:rel-wmd-1}
\text{min}  & \displaystyle\sum\limits_{(w_i,w'_j) \in C} c(w_i,w'_j) T_{i,j} + \displaystyle\sum\limits_{i=1}^{|D|}  c_{max} T_{i,t} & \\
\label{eq:rel-wmd-2}
\text{s.t.:} & \,\, T_{i,t} + \displaystyle\sum\limits_{j | (w_i,w'_j) \in C} T_{i,j}  = D_i   &  i=1 ,\dots, |D| \\
\label{eq:rel-wmd-3}
& T_{t,j}+ \displaystyle\sum\limits_{i | (w_i,w'_j) \in C} T_{i,j}  = D'_j   & j=1,\dots, |D'| \\
\label{eq:rel-wmd-4}
& T_{i,j}, T_{i,t}, T_{t,j} \ge 0 & \text{for all } \quad i,j
\end{align}

We shall note that this formulation has been used before by \cite{pele2009fast} to speed up  EMD \cite{rubner1998metric}
in the context of image retrieval.

\subsubsection{Related Relaxed Word Mover\rq s Distance}
\label{ssec:rel-rwmd}

The Related Relaxed Word Mover \rq s Distance (Rel-RWMD) is a variation of the Rel-WMD, in which we drop constraints of the original formulation in order to obtain a relaxation that can be computed more efficiently.
Rel-RWMD is to Rel-WMD as RWMD is to WMD.

The Rel-RWMD can be computed using  equation
(\ref{eq:rwmd}) with a cost structure given by
(\ref{eq:costs-wmd}).
Let $R'(w_i)$ (resp. $R(w'_j)$ ) be  the set of related words of $w_i$ (resp. $w'_j$) that belong to $D'$ (resp. $D$).
Thus, the Rel-RWMD between documents
$D$ and $D'$ is given by
\begin{equation}
\max \left \{ \sum_{i=1}^{|D|} D_i \min_{w'_j \in R'(w_i)} c(w_i,w'_j),  
\sum_{j=1}^{|D'|} D'_j \min_{w_i \in R(w'_j)} c(w_i,w'_j)  \right \}
\label{eq:rel-rwmd}
\end{equation}
Although not explicit in the above equation,  if $R'(w_i)$ (resp. $R(w'_j))$ is empty then $D_i$ (resp. $D'_j$) is multiplied by $c_{max}$.

To  efficiently evaluate the first term of the maximum in  equation
(\ref{eq:rel-rwmd}) we need to obtain  for each $w$  in $D$ its related words that belong to $D'$, that is, the set $R'(w)$.
By storing  $D'$ as a hash table  we can find them in  $O (r)$ time.
For that, it is enough to traverse the list
of words related to $w$ in cache C and for each word $w'$ in the list we use the hash table of $D'$ to verify whether $w'$ belongs to $D'$. 
Since the second term in the maximum can be calculated analogously we conclude that
the  Rel-RWMD between $D$ and $D'$ can be computed in 
$O((|D| +|D'|)\cdot r)$ time, which is a significant improvement over the $(|D| \cdot |D'|)$ time required
by RWMD  when the size of the documents is considerably larger
than $r$.

Finally, we mention that the
linear time implementation of RWMD presented in Section \ref{ssec:LC-RWMD} can
also benefit from our assumptions.
The first advantage is
that  the matrix
$M$ can be computed faster
since, in order to fill
the row associated with a document
$D$, we just need to consider
the words in the vocabulary 
that are related to $D$ because for the other
words the corresponding entries have value $c_{max}$.
Thus, the addition of the row associated with document $D$ costs $O(|D| \cdot r)$ time rather than the $O(n \cdot |D| \cdot d)$ required by RWMD(L).
The second advantage is the
sparsity of matrix $M$ which allows  handling larger collections/vocabularies.

\section{EXPERIMENTS}
\label{sec:results}

To evaluate our methods, we employ two tasks that involve
the computation of distances between documents.
The first one is the document classification task via $k$-Nearest Neighbors ($k$-NN) that was used to evaluate the WMD algorithm in \cite{kusner2015word}. The second one is the task of identifying related pairs of documents, employed to evaluate
the performance of  paragraph vector  \cite{le2014distributed, dai2015document}.

We compared in terms of test error and computational performance our new distance, Rel-RWMD, against  WMD, RWMD,  Cosine distance and Word Centroid Distance (WCD)
\cite{kusner2015word}. 
The WCD between two documents is given by the Euclidean distance between their centroids,
where the centroid  of a document $D$ is defined as  
$\sum_{i=1}^{|D|} D_i x(w_i)$.
When reporting computational times
we use RWMD(S) and RWMD(L) to distinguish between the
Standard implementation of RWMD   and the one that requires Linear time. Rel-RWMD(S) and  Rel-RWMD(L) are used  analogously. We note that for all RWMD's implementations the preprocessing described at the end of Section \ref{ssec:rwmd} is applied.

Although  we have also implemented/evaluated Rel-WMD,
its results are omitted 
in the next sections since, in general, 
it is competitive with Rel-RWMD in terms of test error
while being much slower.

The methods were  implemented in C++.
The Eigen library \cite{eigenweb}
was used for matrix manipulation and Linear Algebra 
while the OR-Tools library \cite{ortools}
was used for the resolution of flow problems.
All experiments were executed using a single core of an Intel(R) Core(TM) i7-6700 CPU @ 3.40GHz, with 8 GB of RAM.
The code and datasets are available in a GitHub repository \footnote{https://github.com/matwerner/fast-wmd}. 

\subsection{Document classification via $k$-NN}

Our experimental setting follows Kusner et al. \cite{kusner2015word}, where different  distances  are evaluated according to their performance when they are employed by the $k$-NN method to address document classification tasks.

In order to classify a document $D$ from some testing set,
$k$-NN computes the distance of $D$ to each document in
the corresponding training set and then it returns the most frequent class
among the $k$ closest documents to $D$. 
As  stated in \cite{kusner2015word}, motivations for using this evaluation approach, based on $k$-NN , include  its reproducibility and simplicity.

We run  $k$-NN using $k=19$, and, in case of ties,
$k$ is divided by two until there are no more ties.
This setting is slightly different from \cite{kusner2015word}, where $k$ is selected from the set $\{1,3,\ldots,19\}$ based on the lower error rate obtained. 

The parameter $r$ that defines the
number of related words  was selected from the set $S=\{1, 2, 4, \dots, 128\}$ using a $5$-fold cross-validation on top of the training set.  Because we are prioritizing computational performance and, the smaller the $r$ the faster the method, we choose the lowest $r$ whose test error in the cross validation is at most $1\%$ larger than the minimum one found among
all the possibilities in the set $S$.

\subsubsection{Datasets description}
We used the following eight preprocessed datasets \footnote{https://github.com/mkusner/wmd} provided by \cite{kusner2015word}:

\begin{itemize} 
\item \textbf {20NEWS:} Posts on discussion boards for 20 different topics.
\item \textbf {AMAZON:} Product reviews from Amazon for 4 product categories.
\item \textbf {BBCSPORT:} BBC Sport sports section articles for 5 sport between 2004 and 2005.
\item \textbf {CLASSIC:} Sentences from academic works from 4 different publishers.
\item \textbf {OHSUMED:} Medical summaries categorized by different cardiovascular diseases. For computational performance issues, only the first 10 categories of the database were used.
\item  \textbf {RECIPE:} Culinary recipes separated by 15 regions of origin.
\item  \textbf {REUTERS:} News from the Reuters news agency in 1987. The original database contains 90 classes, however, due to problems of imbalance between them, a reduced version with only the 8 most frequent ones was created.
\item \textbf {TWITTER:} Collection of tweets labeled by feelings  ``negative'', ``positive'' and  ``neutral''.
\end{itemize}

For all datasets $70\%$ is used for training and $30\%$ for testing, respecting the partitions provided.
Table \ref{table:dataset-statistics} presents relevant statistics for each of these datasets.

\begin{table}
\begin{center}
\caption{Datasets statistics including training
and testing sets.}
\label{table:dataset-statistics}
\begin{small}
\setlength\tabcolsep{4.0pt} 
\begin{sc}
\begin{tabular}{ lcccc}
\toprule
Name     & \#Docs & $\#$Tokens & \begin{tabular}[x]{@{}c@{}}Avg. tokens\\per Doc\end{tabular} & Classes \\
\midrule
20news   & 18,820 & 22,439 & 69.3  & 20 \\
amazon   & 8,000  & 30,249 & 44.5  & 4\\
bbcsport & 737   & 10,103 & 116.5 & 5 \\
classic  & 7,093  & 18,080 & 38.6  & 4 \\
ohsumed  & 9,152  & 19,954 & 60.2  & 10 \\
recipe   & 4,370  & 5,225  & 48.3  & 15 \\
reuters  & 7,674  & 15,115 & 36.0  & 8 \\
twitter  & 3,108  & 4,489  & 9.9   & 3 \\
\bottomrule
\end{tabular}
\end{sc}
\end{small}
\end{center}
\end{table}

\subsubsection{Results}

Table \ref{table:kusner-error-rate} presents the test errors obtained by the distances under consideration over the eight datasets. We averaged the results for the datasets AMAZON, BBCSPORT, CLASSIC, RECIPE, and TWITTER  following the $5$ predefined train/test splits. The remaining datasets have only one split, and so the average is not necessary. 

Some observations are in order: clearly,  WCD and Cosine 
presented the worst results. Among WMD, RWMD, and Rel-RWMD, there is a balance. The behaviour of WCD and Cosine, as well as the balance between WMD and RWMD, are compatible with the findings/conclusions reached 
in \cite{{kusner2015word}} while 
the  results of Rel-RWMD  suggest that 
our simplifying assumptions work very  well.
The values selected for $r$ ranged from 2 (20NEWS and RECIPE) to 128 (AMAZON) with a median equal 19.5.

\begin{table}
\begin{center}
\caption{Test error (in \%) for different distances and datasets. The datasets with more than one partition had their error rates averaged. The best results are bold faced.} 
\label{table:kusner-error-rate}
\begin{small}
\begin{sc}
\setlength\tabcolsep{4.0pt} 
\begin{tabular}{l|ccccc}
\toprule
Dataset  & Cosine & WCD  & WMD     & RWMD   & Rel-RWMD \\
\midrule
20news   & 30.45 & 36.2 & \bf{24.09} & 24.79 & 25.22 \\
amazon   & 12.90 & 9.04 & 7.21  & \bf{6.87}  &  6.98 \\
bbcsport &  \bf{4.82} & 11.9 & 5.36  & 5.09      & \bf{4.82} \\
classic  &  6.34 & 8.93 & 3.04  & \bf{2.91}  &  3.15 \\
ohsumed  & 45.74 & 47.00 & 42.85 & 43.49     & \bf{41.26} \\
recipe   & 45.71 & 49.20 & 46.56 & 43.63      & \bf{43.20} \\
reuters  &  8.95 & 4.98 & \bf{3.84}  & 3.97  &  4.39 \\
twitter  & 31.97 & 29.4 & 29.14 & \bf{28.95} & \bf{28.95} \\
\midrule
average  & 23.36 & 24.59 & 20.26 & 19.94 & 19.75 \\
\bottomrule
\end{tabular}
\end{sc}
\end{small}
\end{center}
\end{table}

 
Table \ref{table:kusner-computational-cost} presents the  running times in minutes for all distances and datasets examined. 
 First, as expected,  WCD and Cosine are  the fastest distances  since they run in linear time and their preprocessing
phases are very cheap  while WMD is the slowest distance since it has to  solve a transportation problem optimally.
We note that the times of Cosine
were omitted due to the lack of space. 

 It is interesting to examine how  the
distances/implementations related to RWMD perform.
RWMD(S), the original implementation of \cite{kusner2015word}, is the slowest of them
while Rel-RWMD(L) is the  fastest one,
being on average 4.7 times faster than RWMD(L), which is the second fastest.
The main advantage of Rel-RWMD(L)  over RWMD(L) is due
to the time required to build the matrix $M$ since
Rel-RWMD(L) is, on average, 10 times faster.
With regards to the time required to evaluate two
doc's we can also observe a small advantage of Rel-RWMD(L) which is probably  related to the sparsity of $M$.

\begin{table}
\begin{center}
\caption{Computational runtime (in minutes) for different distances and datasets. The datasets with more than one partition have their computational times averaged.} 
\label{table:kusner-computational-cost}
\begin{small}
\setlength\tabcolsep{0.9pt} 
\begin{sc}
\begin{tabular}{l|ccccc}
\toprule
Dataset  & WCD  & WMD & RWMD(S) & RWMD(L) & Rel-RWMD(L) \\
\midrule
20news   & 1.87 & 6,244  & 842  & 68.0 & 13.4 \\
amazon   & 0.30 & 351   & 71.7 & 22.1 & 4.47 \\
bbcsport & 0.01 & 21    & 3.72 & 1.38 & 0.27 \\
classic  & 0.24 & 213   & 45.6 & 10.8 & 2.26 \\
ohsumed  & 0.47 & 1,002  & 158  & 25.1 & 5.20 \\
recipe   & 0.09 & 106   & 27.6 & 2.13 & 0.55 \\
reuters  & 0.27 & 181   & 47.7 & 10.9 & 1.67 \\
twitter  & 0.04 & 3.32  & 1.23 & 0.43 & 0.18 \\
\bottomrule
\end{tabular}
\end{sc}
\end{small}
\end{center}
\end{table}

\vspace{-10pt}

It is important to mention  that the values in Table \ref{table:kusner-computational-cost} do not include  the time required
to estimate the value of $r$. 
In fact, the execution of a 5-fold cross validation on the training set for each potential $r$ incurs a high cost.
However, in practice one can estimate the value of $r$ using a  much smaller set or, even better, set $r$ to a small value, without estimating it, as suggested by the results of Table \ref{table:16-Best}.
This table presents the test errors for $r=1,2,16,128$ and also for the value  estimated via cross validation.  We observe that the test errors remain at the same level, in particular for $r \ge 16$. The running times, though not presented, change very little as expected since the value of $r$ has a small effect in the time complexity of the linear implementation of Rel-RWMD. 

\begin{table}
\begin{center}
\caption{Test Error (in \%) for Rel-RWMD, with $r =1, 2,\ 16,\ 128$ and values estimated via cross-validation, for the different datasets.}
\label{table:16-Best}
\begin{small}
\setlength\tabcolsep{5.5pt} 
\begin{sc}
\begin{tabular}{l|c|cccc}
\toprule
dataset  & Cross Val. & r=1  & r=2      & r=16      & r=128   \\
\midrule
20news   & 25.22  & 25.27 & 25.22     & {\bf 24.90}      & 25.22 \\
amazon   & {\bf 6.98}  & 9.66     & 9.21      & 7.94       & 7.02  \\
bbcsport & 4.82      & 4.00 & \bf{3.64} & 4.91       & 5.55  \\
classic  & \bf{3.15}  &3.62  & 3.56      & 3.20       & 3.18  \\
ohsumed  & {\bf 41.26} &  42.44   & 42.83     & \bf{41.26} & 41.55 \\
recipe   & \bf{43.20} & 43.57 & {\bf 43.20}     & {\bf 43.20}      & 43.52 \\
reuters  & 4.39     & 4.93  & 4.52      & \bf{4.02}  & 4.20  \\
twitter  & 28.95  & 31.52 & 30.60     & \bf{28.91}      & 29.16 \\
\midrule
Average  & 19.75 & 20.63 & 20.35 & 19.79 & 19.92 \\
\bottomrule
\end{tabular}
\end{sc}
\end{small}
\end{center}
\end{table}

\subsection{Identifying related  documents}
For the second task  our experimental setting is inspired on Dai et al. \cite{dai2015document}, where  representations/distances
are evaluated  according to their capacity of recognizing whether a document $D^1$ is more related to a document $D^2$ or to a document $D^3$.  
For achieving this goal,  testing sets are used which contain many triples of documents, namely triplets.
In each triplet, only two  documents are related and  a given distance succeeds if its smallest value is achieved for the related pair.

\subsubsection{Datasets description}

For our experiments, we first downloaded the documents in 
the two testing sets of triplets\footnote{http://cs.stanford.edu/~quocle/triplets-data.tar.gz} provided  in \cite{dai2015document}. The first
set uses papers from Arxiv while the second one uses
articles from Wikipedia. Then, we preprocessed them to remove all non-alphanumeric characters and words contained in a list of stopwords due to its little semantic value. Finally,  to represent the documents we just consider the 
words that have  embeddings  
in the set   that  Google made available.
It is important to note that we are not using the embeddings
of \cite{dai2015document} since they were not provided. 
Table \ref{table:tiplets-dataset-statistics} presents relevant statistics for each of the datasets.

\begin{table}
\begin{center}
\caption{Datasets statistics.}
\label{table:tiplets-dataset-statistics}
\begin{small}
\begin{sc}
\setlength\tabcolsep{4.0pt} 
\begin{tabular}{ lcccc}
\toprule
Name     & \#Docs& \#Triplets & $\#$Tokens & \begin{tabular}[x]{@{}c@{}}Avg. tokens\\per Doc\end{tabular} \\
\midrule
Arxiv            & 47,080 & 19,998 & 260,640 & 1,043.9  \\
Wikipedia        & 58,015 & 19,336 & 415,967 & 429.8 	 \\
\bottomrule
\end{tabular}
\end{sc}
\end{small}
\end{center}
\end{table}

\vspace{-15pt}
\subsubsection{Results}
In this experiment, in contrast to the previous
one,   each document  has its distance evaluated a few times on average,
indeed less than twice. Thus,
building the cache $M$ required for the linear time
implementations of RWMD does not pay off.
In addition, its size would be huge, around
$10^{10}$ entries for Wikipedia
 as an example. 
Therefore, we only executed RWMD(S) and Rel-RWMD(S).

By comparing the statistics of the datasets in Tables \ref{table:dataset-statistics} and \ref{table:tiplets-dataset-statistics}, we  observe that the number of tokens (word embeddings) of the latter is one order of magnitude higher than the former and, as a consequence,  the  preprocessing phase of Rel-RWMD becomes  expensive,  harming the performance gain achieved while computing the distances. Thus,
following our approach, 
we cluster the embeddings before building the cache $C$.
We run $k$-means using a limit of $I=5$ iterations
and setting $k=289 \approx \sqrt{415,967/5}$ for Wikipedia and 
$k=229  \approx \sqrt{260,640/5} $ for Arxiv.
Moreover, motivated by the  discussion/results of the previous 
section we used $r=16$ 
for Rel-RWMD.

Table \ref{table:triplets-accuracy} presents 
 the test errors achieved by the different methods. We can observe a behaviour similar to  the previous task. Once again, both Cosine and WCD achieve the largest test errors  while the others display competitive results.

\begin{table}
\begin{center}
\caption{Test Error (in \%) for different distances and datasets. The best results are bold faced.} 
\label{table:triplets-accuracy}
\begin{small}
\setlength\tabcolsep{3.5pt} 
\begin{sc}
\begin{tabular}{l|ccccc}
\toprule
Dataset          & Cosine & WCD & WMD   & RWMD  & Rel-RWMD \\
\midrule
Arxiv            & 28.83  & 29.99  & {\bf 22.77} & 23.43 & 23.16    \\
Wikipedia        & 27.83  & 29.23 & {\bf 26.74} & 27.01 & 26.90    \\
\bottomrule
\end{tabular}
\end{sc}
\end{small}
\end{center}
\end{table}

The computational times (in minutes) are displayed in Table \ref{table:triplets-computational-cost-alt}. Again WCD and Cosine are the fastest. The former is slower
because it has to compute the centroids of the documents
in its preprocessing phase while the latter does not.
Among the others, Rel-RWMD(S) and RWMD(S), as expected, are much faster than
WMD. For Wikipedia Rel-RWMD(S) is 3 times faster than RWMD(S) while for Arxiv Rel-RWMD(S) it is 27 times faster.

By taking a more in-depth examination of the running times we can also observe that the time consumption of REL-RWMD(S) is highly concentrated on its preprocessing phase when
the cache $C$ is built. Having this structure
available, it computes the distances, on average,
25 and 60 times faster than  RWMD(S)
for Wikipedia and Arxiv, respectively.

\begin{table}
\begin{center}
\caption{Computational runtime (in minutes) for different distances and datasets.} 
\label{table:triplets-computational-cost-alt}
\begin{small}
\setlength\tabcolsep{1.5pt} 
\begin{sc}
\begin{tabular}{l|ccccc}
\toprule
Dataset &     Cosine      & WCD & WMD  & RWMD(S)   & Rel-RWMD(S) \\
\midrule
Arxiv          & 0.01  & 0.18	& 1,996 &	74.8 &  2.72  \\
Wikipedia      & 0.01  & 0.09	& 302  &	11.0	 &  3.36  \\
\bottomrule
\end{tabular}
\end{sc}
\end{small}
\end{center}
\end{table}

\section{CONCLUSION}
\label{sec:conclusions}

In this paper, we presented
an approach to speed up
the computation of WMD and its variants that
relies on the properties
of the distances between embeddings.
The improvements in  time and space complexities 
together with our experimental evaluation provide strong evidence
 that this approach should be employed if one is aiming
 to  compute these distances efficiently.

\ack The  second author is partially supported by CNPq under grant 307572/2017-0 and by FAPERJ, grant Cientista do Nosso Estado E-26/202.823/2018.
\bibliography{references}

\end{document}